\documentclass[10pt, twocolumn, letterpaper]{article}

\usepackage{cvpr}
\usepackage{times}
\usepackage{epsfig}
\usepackage{graphicx}
\usepackage{amsmath}
\usepackage{amssymb}
\usepackage{float}
\usepackage{subfig}
\usepackage{arydshln}
\usepackage{amsthm}
\usepackage{multirow}

\DeclareMathOperator*{\argmin}{arg\!\min}
\DeclareMathOperator*{\argmax}{arg\!\max}

\usepackage[breaklinks=true,bookmarks=false]{hyperref}

\cvprfinalcopy 

\begin{document}

\title{Improved Stereo Matching with Constant Highway Networks\\ and Reflective Confidence Learning}

\author{Amit Shaked and Lior Wolf\\
The School of Computer Science\\
Tel Aviv University\\
{\tt\small amitshaked1@gmail.com, wolf@cs.tau.ac.il}
}

\maketitle

\begin{abstract}
   We present an improved three-step pipeline for the stereo matching problem and introduce multiple novelties at each stage. We propose a new highway network architecture for computing the matching cost at each possible disparity, based on multilevel weighted residual shortcuts, trained with a hybrid loss that supports multilevel comparison of image patches. A novel post-processing step is then introduced, which employs a second deep convolutional neural network for pooling global information from multiple disparities. This network outputs both the image disparity map, which replaces the conventional ``winner takes all'' strategy, and a confidence in the prediction. The confidence score is achieved by training the network with a new technique that we call the reflective loss. Lastly, the learned confidence is employed in order to better detect outliers in the refinement step. The proposed pipeline achieves state of the art accuracy on the largest and most competitive stereo benchmarks, and the learned confidence is shown to outperform all existing alternatives.
\end{abstract}
\section{Introduction}

The modern pipeline for stereo matching, which achieves state of the art results on the most challenging benchmarks, contains a deep neural network for computing the matching score, and a few heuristic post-processing steps. The main goal of these processing steps is to incorporate spatial information in order to verify the plausibility of the proposed matching and to selectively smooth and refine the obtained results.

We present methods that improve the deep matching network by incorporating, among other improvements, a variant of highway networks~\cite{srivastava2015highway} with a multilevel skip-connection structure and a gating signal that is fixed at a constant. Analysis of this architecture is provided, and experimental results are shown to support its advantages over many existing Residual Network alternatives~\cite{residual,dense,resinres,srivastava2015highway}.

To compute the disparity image, a second network is introduced to replace the ``winner takes all'' (WTA) rule, and is currently applied to obtain both the predicted disparity at each position and the confidence in this result as a separate output.  

A confidence measure for the disparity prediction has been the subject of considerable study. More generally, assessing the correctness of a CNN output is also a subject that is under magnifying glass. We propose a new training signal, which we call the {\em reflective loss}. The labels used for this loss are changing dynamically, based on the current success of the CNN on each training sample. In our case, when the disparity predicted by the network for a given example during training is correct, the target confidence label of the sample is 1, otherwise 0. 

The obtained confidence is a crucial part of the subsequent refinement step of the pipeline, in which we detect incorrect disparity predictions and replace them by interpolating neighboring pixels. 

The contributions of this paper are:
(i) In Sec.~\ref{sec:matching} we present a new highway network architecture for patch matching including multilevel constant highway gating and scaling layers that control the receptive field of the network.
(ii) This network is trained with a new hybrid loss (Sec.~\ref{sec:hybrid}) for better use of the description-decision network architecture.
(iii) Computing the disparity image by using a CNN instead of the previously suggested WTA strategy, as depicted in Sec.~\ref{sec:disparity}.
(iv) In Sec.~\ref{sec:reflective} we introduce a novel way to measure the correctness of the output of a CNN via reflective learning that outperforms other techniques in the literature for assessing confidence in stereo matching.
(v) In Sec.~\ref{sec:refinement} we show how to use this confidence score for better outlier detection and correction in the refinement process.
(vi) Achieving the best results on the KITTI 2012 and KITTI 2015 stereo data sets with an error rates of 2.29 and 3.42, respectively, improving the 2.43 and 3.89 of the MC-CNN\cite{newlecun} baseline.
(vii) Improving the fast architecture to achieve the best results on the KITTI 2012 and KITTI 2015 for methods that run under 5 seconds, with an error rate of 2.63 and 3.78 on these benchmarks, in comparison to the 2.82 and 4.62 baseline.
(viii) An open source code\footnote{The code is available at https://github.com/amitshaked/resmatch} for easily using and modifying the pipeline. 

\section{Related work}

Computing the matching cost via convolutional neural networks was firstly introduced by Zbontar and LeCun~\cite{oldlecun,newlecun}. This pipeline was subsequently modified:~\cite{efficient} reduced the computation time with only minor reduction to the accuracy; object knowledge and semantic segmentation were used to create object-category specific disparity proposals~\cite{displets}, and~\cite{adaptive} applied adaptive smoothness constraints using texture and edge information for a dense stereo estimation.

Residual Networks~\cite{residual} (ResNets) are neural networks with skip connections. These networks, which are a specific case of Highway Networks~\cite{srivastava2015highway}, present state of the art results in the most competitive computer vision tasks. However, this is not true with stereo matching. The success of residual networks was attributed to the ability to train very deep networks when employing skip connections~\cite{he2016identity}. A complementary view is presented by~\cite{Veit2016}, who attribute it to the power of ensembles and present an unraveled view of ResNets that depicts ResNets as an ensemble of networks that share weights. 

Very recently, a concurrent tech report proposed a different residual network architecture for the task of image classification, which, like us, employs multilevel skip connection~\cite{resinres}. The two main differences from our architecture are: First, by introducing the learned $\lambda$ coefficient as a constant highway gate, we allow the network to adjust the contribution of the added connections. Second, we add scaling layers to control the receptive field of the network.
Another very recent report entangles the network with more residual connections to create densely connected residual networks~\cite{dense}. We evaluated these architectures and found them (Sec.~\ref{sec:experiments}) to be inferior to the proposed solution, which is much simpler. We also found that adding our constant skip connections contributes significantly to the above architectures.

Estimating the confidence of stereo matches in order to interpolate correspondences is one of the most popular research topics in stereo vision~\cite{gherardi,gehrig,saygili,yoon}. A very recent work~\cite{Seki2016BMVC} was the first to leverage a CNN for stereo confidence measure. They incorporated conventional confidence features to the input of the CNN and trained the network especially for this purpose. Our global disparity network is different in four major ways: (i) We apply a single network to obtain both the confidence score and a much more accurate disparity map. (ii) Our confidence indication is trained with reflective loss that depends not only on the ground truth but also on predicted labels that change dynamically during training. (iii) While \cite{Seki2016BMVC} uses the confidence to improve the performance of the Semi-Global Matching step~\cite{hirschmuller}, we incorporate it in the outlier detection step during the disparity image refinement process. (iv) Our solution is not bounded to stereo matching and our reflective loss is a novel and general technique for evaluating confidence. 

\begin{figure*}[t]
\centering
\includegraphics[height=6cm, width=0.85\linewidth]{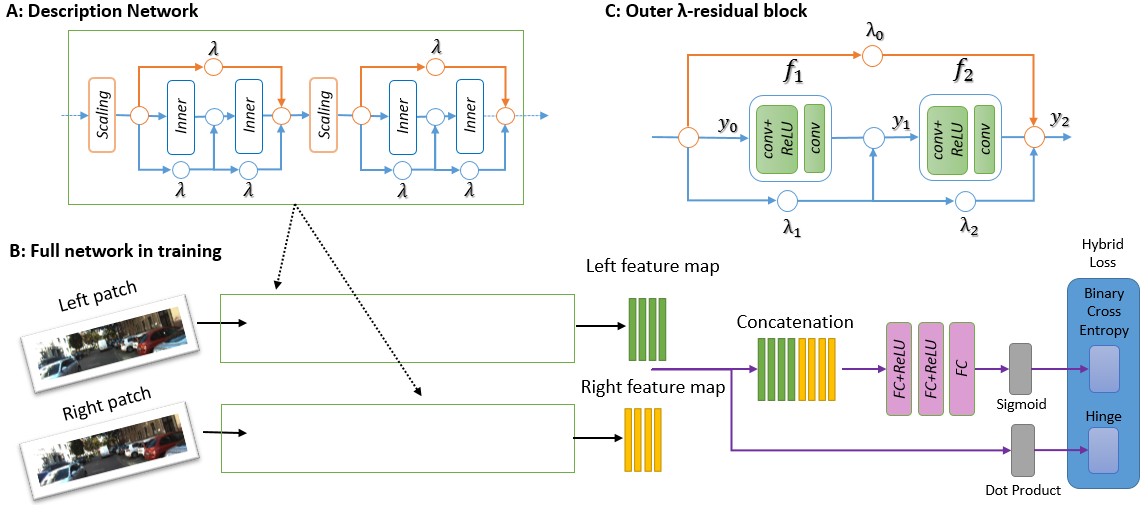}
   \caption{Our $\lambda$-ResMatch architecture of the matching cost network. ReLU activations follow every scaling layer and inner block. (a) The core description network. The inner-$\lambda$-residual blocks are shown in blue and the scaling layers with the outer constant highway skip-connections in orange. (b) Two-tower structure with tied parameters~\cite{matchnet}. The description network outputs the two feature maps, which are the input for two pathways: the first concatenates and passes them to the fully-connected decision network which is trained via the cross-entropy loss, and the second directly employs a Hinge loss criterion to the dot product of the representations. (c) Outer $\lambda$-residual block that consists of two inner $\lambda$-residual blocks.}
\label{fig:matchnet}
\end{figure*}

\section{Computing the matching cost}
\label{sec:matching}
The first step in a modern stereo matching pipeline is based on computing a matching cost at each position for every disparity under consideration. Starting from two images, left and right, for every position $\mathbf{p}$ in the left image and disparity $d$, we compute the matching cost between a patch centered around $\mathbf{p} = (x,y)$ in the left image and a patch centered around $\mathbf{pd} = (x-d, y)$ in the right. The cost is expected to be low for patches centered around the same $3D$ location and high otherwise. Similar to previous work, we employ a convolutional neural network, trained on pairs of small image patches where the true disparity is known.

The novel architecture of our network is presented in Fig.~\ref{fig:matchnet}. It consists of a composition of the following components: A constant highway residual block denoted as inner-$\lambda$-residual, which consists of two convolutional layers with $3\times3$ filters, $1\times1$ padding in order to preserve spatial size, and ReLU non-linearities, followed by a constant highway skip-connection (see Sec.~\ref{sec:lambda}).
Two such blocks, followed by another constant highway connection, are combined into a second level of residual blocks denoted as outer-$\lambda$-residual blocks. Between outer-$\lambda$-residual blocks another convolutional layer, denoted as scaling layer, with $3\times3$ filters and no padding is added, followed by ReLU non-linearity, in order to increase the receptive field of the network. Note that since both inner- and outer-residual blocks maintain the spatial size, the scaling layers are the only factor on the size of the receptive field. For example, a network with 5 scaling layers of $3\times3$ filters has a $11\times 11$ receptive field, no matter how many inner and outer blocks are being used. In this work, we use a description network that is composed of 5 outer blocks, separated by scaling layers.

In order to compare two image patches, two identical (tied weights) description networks are employed and two descriptors vectors are extracted --  $u_{l}$ and $u_r$. During training, two pathways are then used to compare the patches and produce a matching cost. The first pathway, denoted as the decision sub-network, is a fully connected network, which concatenates the two representations into a single vector $[u_l^\top, u_r^\top]^\top$ and is trained via the cross-entropy loss. The second pathway directly employs the hinge loss criterion to the dot product of the two representations $u_l^\top u_r$.

When computing the matching cost of two full size images, the description tensors of the two images, $U^L$ and $U^R$, can be computed in a single forward pass in the description sub-network. Then, for every disparity $d$ under consideration, the matching cost $C(\mathbf{p},d)$ is computed by propagating $U^L(\mathbf{p})$ and $U^R(\mathbf{pd})$ in the decision sub-network. This requires a total of $disparity\_max$ forward passes, which makes it the main factor on the methods runtime.

In order to have the flexibility to choose between accuracy and speed, one applies the full decision network, or uses only the dot-product based similarity score. In both cases, the added term, which is not used in run time, improves the performance of the matching network. This effect and the trade-off is further studied in Sec.~\ref{sec:experiments}.

In the following subsections, we elaborate on each of the structural novelties we introduce to matching networks.

\subsection{Inner-Outer Residual blocks}

Using a deeper network does not always mean a better prediction. For example, Zbontar and LeCun~\cite{newlecun} report that five layer and six layer architectures for the description network are outperformed by the proposed four-layer architecture. Inspired by~\cite{residual}, we want to deepen the network by adding skip connections and employ residual blocks. However, residual networks are ineffective for matching. 

In our experience, stacking residual blocks (with or without constant highway skip connections as described below), leads to great difficulties in making the network converge to a meaningful solution and does not improve the quality of the prediction. We, therefore, suggest further deepening our network by introducing a second level of skip connections and adopting another connection every two inner residual blocks. In addition, spatial pooling layers followed by batch normalization are discarded. They are potentially harmful for matching networks since they reduce the resolution and sensitivity. They are, therefore, incompatible with the stereo matching task. 

We believe that the added capacity of the proposed architecture improves the results both quantitatively and  qualitatively since they allow us to employ new sources of information that are inaccessible in the conventional architectures. For example, our network benefits from the use of color, while the literature reports no added benefit from it (see Appendix~\ref{ap:color}).

\subsection{Constant highway skip connection}
\label{sec:lambda}

In order to further improve the effectiveness of the residual shortcuts, we introduce a constant highway skip connection, in which the identity shortcut of the residual building block is weighted by a learned factor $\lambda$, and formally defined as: 
\begin{equation}
y_{i+1} =  f_{i+1}(y_i) +\lambda_{i+1} \cdot y_i
\end{equation}
In the highway network, the two terms $f_{i+1}(y_i)$ and $y_i$ are weighted by $t_{i+1}$ and $1-t_{i+1}$ respectively, where $t_{i+1}$ is a function of $y_i$. In our case, the weighting collapses to a learned parameter $\lambda_{i+1}$. 

To further understand the effect of the multilevel constant highway connections, we unravel it. Consider an outer block that consists of two inner blocks as shown in Fig.~\ref{fig:matchnet}(c). The formulation of the output $y_2$ is recursive by its nature and can be unrolled as:
\begin{equation}
\label{eq:unrolled}
\begin{split}
y_2 &= \lambda_0 y_0 + \lambda_2 \cdot y_1 + f_2(y_1) \\
	&= \lambda_0 y_0 + \lambda_2 \big( \lambda_1 y_0 + f_1(y_0) \big) +f_2\big(\lambda_1 y_0 + f_1(y_0)\big) \\
    &= (\lambda_0 + \lambda_2 \lambda_1)y_0 + \lambda_2 f_1(y_0) +f_2\big(\lambda_1 y_0 + f_1(y_0)\big)
\end{split}
\end{equation}
One can see that the added parameter $\lambda_2$ controls the flow of $f_1$, $\lambda_1$ balances the input to $f_2$ and the outer parameter $\lambda_0$ controls $y_0$. That way, when the residual network is interpenetrated as an ensemble of the possible paths~\cite{Veit2016}, the learned parameters determine the contribution of each path to the ensemble. For example, we observed that they adopt much smaller values in the upper layers of the network to reduce the effect of the shortest paths and bias the network toward deeper representations (see Appendix~\ref{ap:lambda}). 

All $\lambda$-parameters are initialized with the value of one in order to emulate vanilla residual connections, and are then adjusted by back-propagation, as usual. No regularization is applied to this term.

\subsection{Hybrid loss}
\label{sec:hybrid}

While further processing the output of the two descriptor networks improves the ability to discriminate between matching and non-matching patches, it comes at the cost of making the descriptors less explicit. We, therefore, suggest combining two losses together: a hinge loss over the dot product $s=u_l^\top u_r$ and the cross-entropy over the decision network's output $v$. Similar to~\cite{newlecun} we consider pairs of examples, matching and non-matching, centered around the same image position, and the compound loss is given by: 
\begin{equation}
loss = \alpha\cdot \mathit{XEnt}(v_+,v_-) + (1-\alpha)\cdot Hinge(s_+,s_-)
\end{equation}
where $Hinge(s_+,s_-) = max(0,m+s_- -s_+)$ and $\mathit{XEnt}(v+,v_-)= -(\log(v_-) + \log(1-v_+)\big)$

Note that the dot product produces similarity score. Therefore, when choosing the fast pathway the output is multiplied by $-1$ to represent the matching cost. A margin of $m=0.2$ for the hinge loss and $\alpha = 0.8$ are used throughout the experiments.

\section{Computing the disparity image}
\label{sec:disparity}

The computation of the matching cost results in a map $C(\mathbf{p}, d)$ of size $H\times W\times disparity\_max$ in which the matching cost of every position is computed for every possible disparity. The goal of the next stage is to output the disparity image $D(\mathbf{p})$ of size $H\times W$ with the predicted disparity in every position.

Modern stereo matching pipelines use few post-processing steps, and then apply the ``winner takes all'' strategy: 
$D(\mathbf{p}) = \argmin_d C(\mathbf{p}, d)$. The post processing is required, since even with improved matching networks, in order to be competitive, the method needs to incorporate information from the neighboring pixels that is beyond a simple maximization. Following Mei et al~\cite{mei}, we begin by applying cross-based cost aggregation~\cite{cbca} (CBCA) to combine information from neighboring pixels by averaging the cost with respect to depth discontinuities, continue with semi global matching~\cite{hirschmuller} (SGM) to enforce smoothness constraints on the disparity image, and then apply few more iterations of the cost aggregation, as described in~\cite{newlecun}.

\begin{figure}[t]
\centering

\subfloat[Reference image]{%
  \includegraphics[width=8cm]{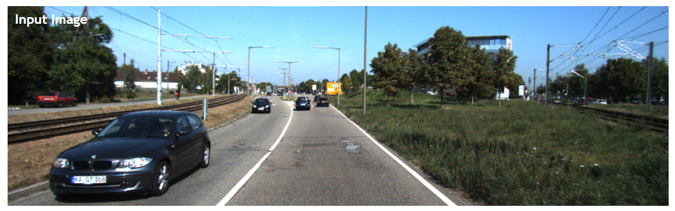}%
}

\subfloat[Prediction errors before applying the disparity network]{%
  \includegraphics[width=8cm]{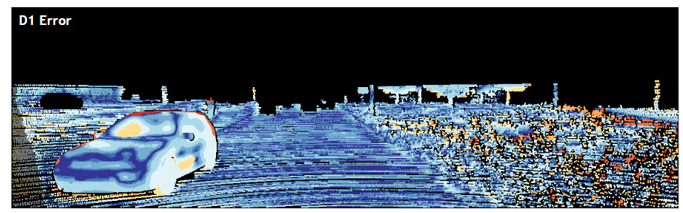}%
}

\subfloat[Prediction errors after applying the disparity network]{%
  \includegraphics[width=8cm]{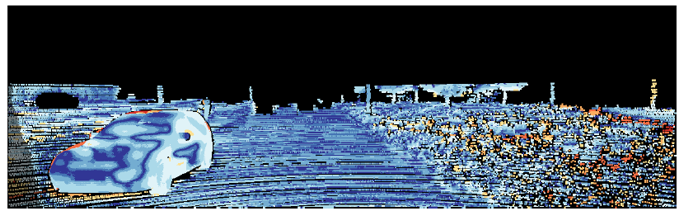}%
}
\caption{An example taken from KITTI 2015 data set showing the effect of the disparity network. Observe the errors in the predicted car glass disparities before and after applying the disparity network. }
\label{fig:kittiexample}
\end{figure}

While CBCA and SGM contribute greatly to the success of modern stereo matching pipelines, they are limited and error especially in challenging situations where machine learning can help. These situations include occluded or distorted areas, highly reflective or sparse texture regions and illumination changes. 
Fig.~\ref{fig:kittiexample}(b) presents one example in which these schemes fail to correct the matching in reflective regions such as car glass. 

To overcome this challenge, while one can follow the footsteps of Guney and Geiger\cite{displets} 
and use object-category specific disparity proposals, this requires explicit object knowledge and semantic segmentation, which we choose to avoid for two reasons: the introduced computational complexity, and the loss of generality associated with specific objects models. Instead, we propose to apply a learned criterion and replace the WTA approach. We construct a global disparity convolutional neural network and propagate the entire matching cost map to output the disparity at each position. An example of how this method can help in challenging situations is presented in Fig.~\ref{fig:kittiexample}(c). This network is trained with a novel reflective loss to simultaneously produce a confidence measure in the network's disparity prediction, to be used later in the refinement process, as described in Sec~\ref{sec:refinement}.

\subsection{Global Disparity Network}

The data set used to train the disparity network is composed of processed images from the matching cost network training data. For each pair of left and right images, we compute the full image size matching cost at each possible disparity as described in Sec.~\ref{sec:matching}, and then apply CBCA and SGM. Note that the matching network returns independent probability estimations, and that post CBCA and SGM, the values can become negative and not bounded to a specific range. We, therefore, apply $Tanh$ in order to bring the values to the fixed range $[-1,1]$. The target (ground truth) value for each matching costs patch is the disparity of its central pixel. Sampling $9\times 9$ patches this way, we collect 25 (17) million training examples for training the KITTI 2012 (2015) disparity network.

\begin{figure*}[t]
\centering
\includegraphics[height=4cm, width=\linewidth]{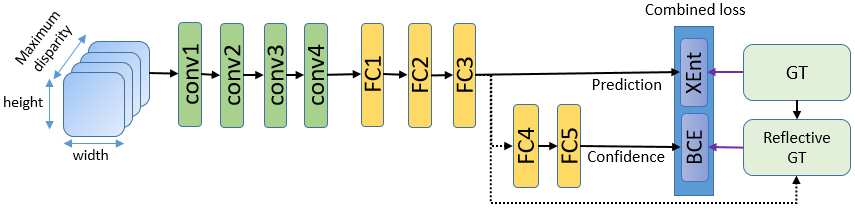}
   \caption{The Global disparity network model for representing disparity patches. ReLU units are used as activation functions following every convolution and fully connected layer. Two layers are considered as target layers: FC3 on which LogSoftMax is applied to determine the predicted disparities, and FC5 which depicts the confidence measure. Cross-entropy loss for the prediction and binary cross-entropy loss for the confidence measure are combined together using 85:15 weights, respectively.}
\label{fig:dispnet}
\end{figure*}

The patches are fed to the global disparity network (GDN) as described in Fig.~\ref{fig:dispnet}. Two layers are considered as target layers: FC3, that outputs the vector $\mathbf{y}$, which is the score $y_i$ for every disparity $d_i$, and FC5, which depicts the confidence $c$ in the prediction. The loss function on FC3, inspired by~\cite{efficient}, is a weighted cross-entropy loss that centers the mass around the ground truth $y^{GT}$, with respect to the error metric of the data set: 
\begin{equation}
loss(\mathbf{y}, y^{GT}) = -\sum_{y_i}{p(y_i, y^{GT})\cdot \log{\frac{e^{-y_i}}{\sum_j{e^{y_j}}}}}
\end{equation}
where $p(y_i, y^{GT})$ is a smooth target distribution, centered around the ground-truth $y^{GT}$. 
For the KITTI data set, we are interested in 3-pixel error metric and use:
\begin{equation}
p(y_i, y^{GT}) =
\left\{
	\begin{array}{ll}
		 \lambda_1  & \mbox{if } |y_i - y^{GT}| \leq 1 \\
		 \lambda_2  & \mbox{if } 1 < |y_i - y^{GT}| \leq 2 \\
         \lambda_3  & \mbox{if } 2 < |y_i - y^{GT}| \leq 3 \\
         0  & \text{otherwise} \\
	\end{array}
\right.
\end{equation}
The main difference from~\cite{efficient} is that we allow real-valued $y^{GT}$ and the loss is modified accordingly. The values used in our work are $\lambda_1 = 0.65$, $\lambda_2 = 0.25$, $\lambda_3 = 0.1$. A model that outputs the disparity instead of the scores vector was also tested with different regression losses and found to be inferior to our model.

\subsection{Reflective confidence}
\label{sec:reflective}

In order to obtain a confidence measure from the disparity network, we simultaneously train a binary classifier consists of two fully-connected layers via the binary cross-entropy loss. The training labels for this loss reflect the correctness of the score vector \textbf{y}, which is also the input to the classifier, as shown in Fig~\ref{fig:dispnet}. After each forward pass, $\argmax_i{y_i}$ is compared to the ground truth disparity $y^{GT}$. If the prediction is correct, i.e differs from the ground truth by less than one pixel, the sample is considered positive, otherwise negative. Note that although the KITTI data set requires an error less than three pixels, we notice that training the confidence allowing a three pixel error (and not just one) causes too many positive samples and is not effective. 

This loss is unconventional in the sense that the target value depends not only on the ground truth, but also on the activations of the network. To our knowledge, this is the first loss in the literature that is based on labels that change dynamically during training in this way. 

This reflective loss is combined with the weighted cross-entropy loss of FC3 using 15:85 weights, respectively. We employ mini-batches of size 128 and a momentum of 0.9. The network was trained for 15 epochs, starting with a learning rate of 0.003 and decimating it on the 12th epoch.

\section{Disparity refinement}
\label{sec:refinement}
While the global disparity network greatly contributes to the quality of the predicted disparity image, it can still suffer from known issues such as depth discontinuities and outlier pixel predictions. The goal of the third and last stage is to refine the disparity image and output the final prediction. Similar to~\cite{newlecun,mei}, we employ a three-step refinement process: (i) left-right consistency check for outlier pixel detection and interpolation, in which we incorporate our confidence score; (ii) sub-pixel enhancement in order to increase the image resolution, and (iii) median and bilateral filtering for smoothing the disparity image without blurring the edges. The second and third steps are performed exactly as in~\cite{newlecun}. The first step is described below.

Denote $C^{L}(\mathbf{p})$ as the confidence score at position $\mathbf{p}$ of the prediction $d = D^{L}(\mathbf{p})$ obtained by using the left image as a reference, and  $C^{R}(\mathbf{pd})$ the confidence score at the correspondent position $\mathbf{p-d}$ of the prediction $D^{R}(\mathbf{pd})$, obtained by using the right image as a reference.
We label each position $\mathbf{p}$ applying these rules in turn:
\begin{align*}
 \textit{correct}  & \quad \mbox{if} \quad |d - D^{R}(\mathbf{pd})| \leq \tau_1  \quad \mbox{or} \\
  & \quad  \big(C^{L}(\mathbf{p}) \geq \tau_2 \; \mbox{and} \; C^{L}(\mathbf{p}) - C^{R}(\mathbf{pd}) \geq \tau_3 \big) \\
   \textit{mismatch} & \quad \mbox{if there exist $\hat{d} \neq d$ s.t. } \; |\hat{d} - D^{R}(\mathbf{p\hat{d}})| \leq \tau_{4} \\
   \textit{occlusion} & \quad \text{otherwise}
\end{align*} 
That means a pixel is labeled as \textit{correct} if the two predictions $D^{L}(\mathbf{p})$ and $D^{R}(\mathbf{pd})$ match, or they don't match but the reference prediction $D(\mathbf{p})=D^L(\mathbf{p})$ is much more reliable.
When neither holds, a pixel is considered  \textit{mismatch} if there exist another disparity $\hat{d}$ such that if it were the prediction, it would have matched $D^{R}(\mathbf{p\hat{d}})$. If none exist, the pixel is considered as \textit{occlusion}.
Throughout our experiments, we use $\tau_1=1$ the maximum left-right prediction disagreement, $\tau_2=0.7$ the minimum confidence score in the prediction, $\tau_3=0.1$ the minimum left-right confidence gap, and $\tau_4=1$ the maximum left-right prediction disagreement for other possible disparities.

For pixels labeled as \textit{mismatch}, we want the disparity value to come from the reliable neighboring pixels and so take the median of the nearest neighbors labeled as \textit{correct} from 16 different directions. The value of outliers $\mathbf{p}$ marked as \textit{occlusion} most likely come from the background. Therefore, the interpolation is done by moving left until the first \textit{correct} pixel and use its value.

\section{Experimental results}
\label{sec:experiments}
We evaluated our pipeline on the three largest and most competitive stereo data sets: KITTI 2012, KITTI 2015 and Middlebury. Comparisons with the state of the art and components analysis are provided.

\subsection{Benchmark results}
\textbf{KITTI stereo data sets}:
The KITTI 2012~\cite{kitti2012} data set contains 194 training and 195 testing images, and the KITTI 2015~\cite{kitti2015} data set contains 200 training and 200 testing images. The error is measured as the percentage of pixels for which the true disparity and the predicted disparity differ by more than three pixels. The leader-boards of the two data sets are presented in Tab.~\ref{tab:kitti2015benchmark} and Tab.~\ref{tab:kittibenchmark}. The reported error rates were obtained by submitting the generated disparity maps of the testing images to the online evaluation servers. Our accurate method is ranked first on both benchmarks and improved the error rate of the mc-cnn~\cite{newlecun} baseline from 2.43 to 2.29 on KITTI 2012 and from 3.89 to 3.42 on KITTI 2015.

Our fast architecture was also submitted to the online servers, and a comparison between methods that run under five seconds is presented in Tab.~\ref{tab:fast_kitti2015} and Tab.~\ref{tab:fast_kitti}.

The runtime was measured by testing our pipeline on the NVIDIA Titan X (pascal) graphics processor unit. 

\begin{table}[t]
\begin{center}
\begin{tabular}{|l l l c c c|}
\hline
		 	& Method 				& Set.	& NOC 		& ALL 		& runtime  \\ \hline
		\bf{1}	& \bf{Ours}  		&   		&\bf{2.91} 	&\bf{3.42} 	& \bf{48s} \\
        2 		& Displets v2\cite{displets} & S 	& 3.09 		& 3.43 	& 265s \\
        3 		& PCBP\cite{ntouskos2016confidence}  				& 		& 3.17 		& 3.61 	& 68s \\
        \bf{4} & \bf{Ours-fast} & & \bf{3.29} & \bf{3.78} & \bf{2.8s} \\
        5 		& MC-CNN-acrt\cite{newlecun} 			& 		& 3.33 		& 3.89 	& 67s \\		
\hline
\end{tabular}
\end{center}
\caption{The highest ranking methods on KITTI 2015 due to November 2016, ordered by the error rate for all pixels. The S in the settings indicates the use of semantic segmentation. Very recently, two more anonymous submissions were submitted to the online evaluation server. CNNF+SGM achieves an error rate of 3.60 for all pixels and 3.04 for non-occluded pixels, and SN that achieves 3.66 and 3.09, respectively. We do not know whether or not they use segmentation.}
\label{tab:kitti2015benchmark}
\end{table}

\begin{table}[t]
\begin{center}
\begin{tabular}{|l l l c c c|}
\hline
		 		& Method 				& Set 		& NOC 		& ALL 		& runtime  \\\hline
	\bf{1} 		& \bf{Ours}  		&			& \bf{2.29}	& \bf{3.36} & \bf{48s} \\
		2 		& PCBP\cite{ntouskos2016confidence}  				&  			& 2.36 		& 3.45 		& 68s \\
		3 		& Displets v2\cite{displets}  			& S 		& 2.37 		& 3.09 		& 265s \\
		4 		& MC-CNN-acrt\cite{newlecun} 			&  			& 2.43 		& 3.63 		& 67s \\
		5 & cfusion\cite{ntouskos2016confidence}	&  MV	& 2.46 		& 2.69 		& 70s \\
\hline
\end{tabular}
\end{center}
\caption{The highest ranking methods on KITTI 2012 due to November 2016, ordered by the error rate for non occluded pixels. The S in the settings indicates the use of semantic segmentation and MV the use of more than two temporally adjacent images. The very recent anonymous submissions mentioned in Tab.~\ref{tab:kitti2015benchmark} were also submitted here, where CNNF+SGM achieves 2.28 error rate for non occluded pixels and 3.48 for all pixels, and SN 2.29 and 3.50 respectively.}
\label{tab:kittibenchmark}
\end{table}

\begin{table}[t]
\begin{center}
\begin{tabular}{|c c c c c|}
\hline
Rank & Method & NOC & ALL & runtime  \\
\hline
\bf{1} & \bf{Ours-fast} & \bf{3.29} & \bf{3.78} & \bf{2.8s} \\
2& DispNetC\cite{dispnet} 		&4.05  &4.34  & 0.06s \\
3& Content-CNN\cite{efficient}  	&4.00  &4.54 & 1s \\
4& MC-CNN-fast\cite{newlecun}  	& ?  & 4.62 & 0.8s \\
5& SGM+CNN(anon)	& 4.36 & 5.04 & 2s \\
\hline
\end{tabular}
\end{center}
\caption{The highest ranking methods on KITTI 2015 for methods under 5 seconds due to November 2016.}
\label{tab:fast_kitti2015}
\end{table}

\begin{table}[t]
\begin{center}
\begin{tabular}{|c c c c c|}
\hline
Rank & Method & NOC & ALL & runtime  \\
\hline
\bf{1} & \bf{Ours-fast} &  \bf{2.63} & \bf{3.68} & \bf{2.8s} \\
2 & MC-CNN-fast\cite{newlecun} & 2.82 & ? & 0.7s \\
3 & Content-CNN\cite{efficient} &  3.07 & 4.29 & 0.7s \\
4 & Deep Embed\cite{deep_embed}  &  3.10 & 4.24 & 3s \\ 
5 & SPS-st\cite{Yamaguchi14}  &  3.39 & 4.41 & 2s \\ 
\hline
\end{tabular}
\end{center}
\caption{The highest ranking methods on KITTI 2012 for methods under 5 seconds due to November 2016.}
\label{tab:fast_kitti}
\end{table}

\textbf{Middlebury stereo data set}:
The Middlebury stereo data set contains five separate works in the years 2001~\cite{mb2002}, 2003~\cite{mb2003}, 2005~\cite{mbpal2007}, 2006~\cite{mb2007}, and 2014~\cite{mb2014}. The image pairs are indoor scenes given in a full, half and quarter resolution, and rectified perfectly using precise 2D correspondences for
perfect rectification, or imperfectly using standard calibration procedures. We trained our network on half resolution, due to the limited size of our GPU's memory card, on pairs rectified imperfectly, since only two out of fifteen image pairs in the test set are rectified perfectly. The error is measured for pixel disparity predictions that differ from the ground truth by more than two pixels, and is always computed on full resolution. Hence, when training on half resolution, we are interested in less than one pixel error.
The data set contains 60 image pairs for which the ground truth is available, but unlike KITTI, the maximal disparity is not fixed and varies between 30 and 800.
Our global disparity network feature input plane is the size of the maximal disparity, and since there are only a few pairs for each possible value, there was not enough data to train the network at a fixed size. We have, therefore, tested our $\lambda$-ResMatch architecture using the post processing of~\cite{newlecun}. As can be seen in Tab~\ref{tab:main_results}, the fast architecture introduces substantial improvement, lowering the validation error from 9.87 reported on \cite{newlecun} to 9.08. We were not able to reproduce the 7.91 error rate reported for the accurate architecture. Training MC-CNN~\cite{newlecun} with its published code obtained 8.18 validation error that we improved with $\lambda$-ResMatch to 8.01. 

\subsection{Components Analysis}
\label{ablation}
In order to demonstrate the effectiveness of our novelties in each stage of the pipeline, we gradually tested them on the above data sets. Table~\ref{tab:main_results} reports the contribution of adding each stage. One can see that on the KITTI 2015 data set, the greatest improvement comes from employing the global disparity network, while on KITTI 2012 it is the novel constant highway network. This is due to the fact that vehicles in motion are densely labeled in KITTI 2015 and car glass is included in the evaluation, and, therefore, reflective regions are more common.

\begin{table}[t]
\begin{center}
\begin{tabular}{|lccccc|}
\hline
                         & \multicolumn{2}{l}{KITTI 2012} & \multicolumn{2}{l}{KITTI 2015} & MB \\ \cline{2-3} \cline{4-6} 
\textbf{Component}			& fast          & act			& fast	& act   	& fast    \\ \hline
mc-cnn~\cite{newlecun}		& 3.02          & 2.61         	& 3.99  & 3.25  	& 9.87	       \\
$\lambda$-ResMatch			& 2.73          & 2.45         	& 3.69  & 3.15   	& 9.08	\\
GDN 						& 2.66          & 2.40         	& 3.18  & 2.87    	& --		\\
CR       					& 2.65          & 2.38     		& 3.16   & 2.83   	& --		\\ \hline
\end{tabular}
\end{center}
\caption{Validation error on KITTI and Middlebury stereo data sets. The error is computed by splitting the data into 80-20 train validation. The $\lambda$-ResMatch row represents the error when applying our constant highway architecture with the post processing of~\cite{newlecun}. The GDN row stands for the error after replacing the WTA rule with the global disparity network, and the CR is the error after incorporating our confidence measure in the refinement process.}
\label{tab:main_results}
\end{table}

\textbf{$\lambda$-ResMatch architecture}:
We have tested our accurate architecture for the matching cost network thoroughly and compared it to five different architectures: (i) the baseline for our work MC-CNN~\cite{newlecun}, (ii) conventional highway network, (iii)  ResNets~\cite{residual}, (iv) the concurrent work of Densely Connected Residual network~\cite{dense}, and (v) the concurrent work of Residual networks of residual networks~\cite{resinres} which also suggests to add another level of residual connection. In early experiments, we used the published code of these architectures as our matching cost network, but the results were far from competitive. We then removed the batch normalization and pooling layers and replaced them with our scaling layers, except the work of densely connected in which we used their original ``transition layers''. These results are reported in Tab.~\ref{tab:res_compare}, at the first row of every architecture. The second row contains further experiments of other architecture variants where we replaced the vanilla residual shortcuts with our constant highway skip connections. The results show that the multilevel constant connections contribute in almost all cases.
To let the comparison be as direct as possible, we tested the accurate architecture with and without hybrid loss training. One can see that the $\lambda$-ResMatch architecture achieves the best accuracy in the task of stereo matching on all data sets, and that the added hybrid loss further improves the results.
\begin{table}[t]
\centering
\scalebox{0.9}{
\begin{tabular}{|c||c|c||c|c|c|}
\hline
		& Inner 	& Outer & KITTI & KITTI & MB \\
		& shortcut 	& shortcut & 2012 & 2015 &  \\ \hline
        
{mc-cnn\cite{newlecun}}      
		& -				& -					& 2.84		& 3.53   	 & 9.73                           \\ \hline
{Highway\cite{srivastava2015highway}}      
		& -				& -					& 2.81		& 3.51   	 & 9.77                           \\ \hline

{ResNet\cite{residual}}                           
		& A 			& -					& 2.82		& 3.71		&  10.03                   \\ \cline{2-6} \hdashline
$\lambda$ variant        & $\lambda$ 	& -					& 2.81		& 3.55		& 10.01                    \\ \hline
{DC\cite{dense}}                           
		& A 			& -					& 3.86			& 5.02		&  11.13                       \\ \cline{2-6} \hdashline
$\lambda$ variant        & $\lambda$ 	& -					& 3.42			&  4.43    	 	&  11.07                        \\ \hline
{RoR\cite{resinres}}      
		& A				& C					& 2.86		& 3.52   	&  9.68                      \\ \cline{2-6} \hdashline
\multirow{1}{*}{$\lambda$ variant}	& $\lambda$		&$\lambda\cdot$ C 	& 2.84		& 3.53   	&  9.95                      \\ \hline

\multirow{1}{*}{Variants of} 
		& A				& A					& 2.78		& 3.49   	 & 9.63                          \\ \cline{2-6}
\multirow{1}{*}{ our method}                                
        & $\lambda$		& A					& 2.75		& 3.42    	 & 9.83                         \\ \cline{2-6}
\multirow{1}{*}{without the } 
        & A				& $\lambda$			& 2.78		& 3.46   	 & 10.3                           \\ \cline{2-6}
\multirow{1}{*}{hybrid loss} 
		& $\lambda$		& $\lambda$			&2.73		& 3.42   &  9.60                       \\ \hline
\multirow{1}{*}{\textbf{$\lambda$-ResMatch}}     
		& $\lambda$		& $\lambda$			& \textbf{2.71}		& \textbf{3.35}   & \textbf{9.53}     \\ \hline
\end{tabular}}
\caption{ The validation errors of different architectures and their $\lambda$-variants, when trained on 20\% of the data. ``\textbf{A}'' shortcut is the identity connection, ``\textbf{C}'' is $1X1$-convolution and ``$\mathbf{\lambda}$'' is our constant highway skip-connection.}
\label{tab:res_compare}
\end{table}

\textbf{Reflective confidence}:
To evaluate the performance of our new method for confidence indication, we compared it to the six most widely used techniques, using the AUC measure. These techniques belong to different categories according to which aspects of stereo cost estimation they take into account \cite{confpaper}. The notations we use to describe the different methods are: 
$d_1(\mathbf{p}) =  D(\mathbf{p}) = D^L(\mathbf{p})$ the predicted disparity at position $\mathbf{p}$ when using the left image as a reference, $c_1(\mathbf{p}) = C_{SGM}(\mathbf{p},d1)$ the matching cost of the prediction disparity (before applying the global disparity network), $c_2$ the second smallest local minimum, and $prob(\mathbf{p}) = C_{GDN}(\mathbf{p},d_1) = \max_d{C_{GDN}(\mathbf{p},d)}$ the probability of the predicted disparity.

The evaluated methods are (i) Matching Score Measure (MSM)\cite{msm} which assigns higher confidence to lower costs: $C_{MSM} = -c_1$. (ii) The probability (PROB) of the prediction after applying the disparity network $C_{prob}=prob$, (iii) The Curvature (CUR) of the matching cost $C_{CUR}=-2\cdot c(d_1) + c(d_1-1) + c(d_1+1)$ that is widely used in the literature, (iv) the Peak Ratio (PKRN)\cite{confpaper} is the ratio between the second smallest cost and the smallest $C_{PKRN}=\frac{c_2}{c_1}$, (v) the Negative Entropy Measure (NEM)\cite{entropy} $p(d) = \frac{e^{-c_1}}{\sum_{d}{e^{-c(d)}}}$, $C_{NEM}=-\sum_{d}{p(d)\log{p(d)}}$ that takes into consideration the entire cost curvature, and (vi) the Left Right Difference (LRD)\cite{confpaper} which utilizes both left-right consistency check and the margin between the two smallest minima of the cost: $C_{LRD}=\frac{c_2-c_1}{|c_1-\min{c_R(x-d_1,y,d_R)}|}$.

We tested the above measures on 40 random validation images from the KITTI 2012 and the KITTI 2015 data sets. The results for KITTI 2015 presented in Fig~\ref{fig:kitti2015_conf}, and the very similar results for KITTI 2012 that can be found in Appendix~\ref{ap:conf}, show that our reflective confidence measure performs better on almost every image. The average score over the entire data set in Tab.~\ref{tab:confidence} shows that it is also the overall most accurate in both data sets.

\begin{table}[t]
\centering
\scalebox{0.8}{
\begin{tabular}{|l|l|l|l|l|l|l|l|}
\hline
                & Ref 				& MSM 	&Prob	& CUR	&PKRN	&NEM	&LRD   \\ \hline
KITTI2012       & \textbf{0.943}    &0.928	&0.648	& 0.772	&0.930	&9.190	&0.833	\\ \hline
KITTI2015 		& \textbf{0.894} 	&0.850  &0.758	& 0.832	&0.853	&0.864	&0.812  \\ \hline 
\end{tabular}}
\caption{The average AUC over the entire validation set for different confidence measures.}
\label{tab:confidence}
\end{table}

\begin{figure}[t]
\centering
  \includegraphics[width=8cm, height=5cm]{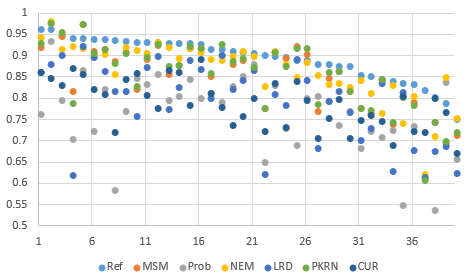}%
\caption{AUC of confidence measures on 40 random validation images from the KITTI 2015 stereo data set.}
\label{fig:kitti2015_conf}
\end{figure}

\section{Discussion}

It is interesting to note that unlike the most recent state of the art results, we make no use of semantic segmentation. Semantic segmentation employs additional training data that is not used by our method and requires an additional runtime. Nevertheless, it would be interesting to see whether the benefits of semantic segmentation are complementary to the increased performance we obtain, in which case, our results could be further improved with little effort.

The need for specific residual architectures for matching probably goes hand in hand with the favorable performance of moderate depth networks, the detrimental effect of batch normalization, and other unique practices in this domain. More study is required in order to understand what sets this problem apart from categorization problems.

We believe that the reflective loss can be extended both to other problems and to other applications such as the gradual learning schemes self-paced learning~\cite{kumar2010self} and curriculum learning \cite{bengio2009curriculum}. 

\section*{Acknowledgments}
This research is supported by the Intel Collaborative Research Institute for Computational Intelligence (ICRI-CI).

{\small
\bibliographystyle{ieee}
\bibliography{stereopipeline}
}

\appendix
\section{The benefit of color}
\label{ap:color}

\begin{figure*}[t]
\centering

\subfloat[Reference image]{%
  \includegraphics[width=8cm]{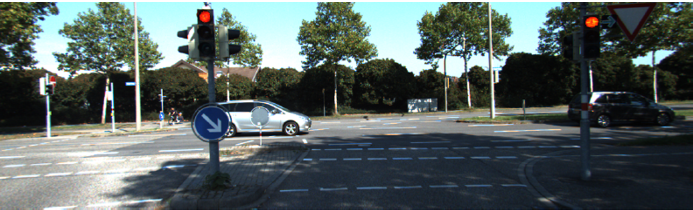}%
}

\subfloat[Disparity map prediction and its errors before incorporating color information]{%
  \includegraphics[width=0.4\linewidth]{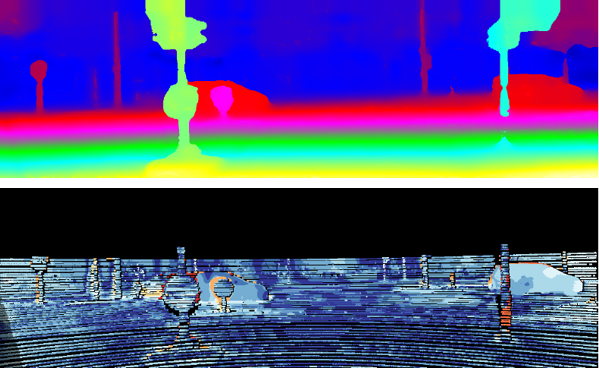}%
} \qquad
\subfloat[Disparity map prediction and its errors after incorporating color information]{%
  \includegraphics[width=0.4\linewidth]{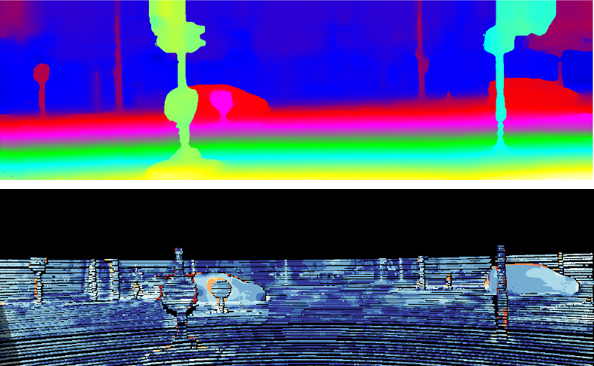}%
}
\caption{An example taken from KITTI 2015 data set showing the effect of color in the $\lambda$-ResMatch architecture. Observe the errors where the color differences between the traffic light and the background are very delicate.}
\label{fig:colorexample}
\end{figure*}

Previous architectures in the literature for computing the matching cost report no benefit from using color information~\cite{newlecun,efficient}. In our experiments we observed that after deepening our network, the use of the three input channels contributes to the accuracy of the disparity prediction, especially around areas of delicate color differences between the object and its background. An example for this phenomenon is shown in Fig.~\ref{fig:colorexample}, and the average improvements over the validation sets of KITTI and Middlebury are presented in Tab.~\ref{tab:color_compare}.

\begin{table*}[t]
\begin{center}
\begin{tabular}{|lccccc|}
\hline
		& \multicolumn{2}{c}{KITTI 2012} & \multicolumn{2}{c}{KITTI 2015} & MB \\ \cline{2-3} \cline{4-6} 
								& Fast    & Accurate & Fast	& Accurate	& Fast    \\ \hline
mc-cnn~\cite{newlecun}			& 3.02    & 2.61    & 3.99  & 3.25  	& 9.87	       \\
mc-cnn~\cite{newlecun}+color	& 3.02    & 2.61    & 3.99  & 3.25  	& 9.87	\\ \hdashline
$\lambda$-ResMatch (no color)	& 2.82    & 2.51    & 3.79  & 3.18   	& 9.35	\\
$\lambda$-ResMatch				& 2.73    & 2.45    & 3.69	& 3.15  	& 9.08 \\ \hline
\end{tabular}
\end{center}
\caption{The benefits from color use in $\lambda$-ResMatch and the baseline MC-CNN. The errors reported are the validation errors after applying the post processing steps used in \cite{newlecun}.}
\label{tab:color_compare}
\end{table*}

\section{Multilevel constant highway connections}
\label{ap:lambda}
We study the effect of the added inner and outer constant highway connections. The amount of the input that is added to each outer block, according to Eq.~\ref{eq:unrolled}, is determined by $\lambda_0+(\lambda_1 \cdot \lambda_2)$. When the network is interpreted as an ensemble of the possible paths~\cite{Veit2016}, the value of $\lambda_{i,0}+(\lambda_{i,1} \cdot \lambda_{i,2})$ for outer-block $i$ determines the contribution of the sub-network that consists of $i$ outer-blocks to the ensemble. 

\begin{figure*}[t]
\centering
  \includegraphics[width=0.9\linewidth, height=7.5cm]{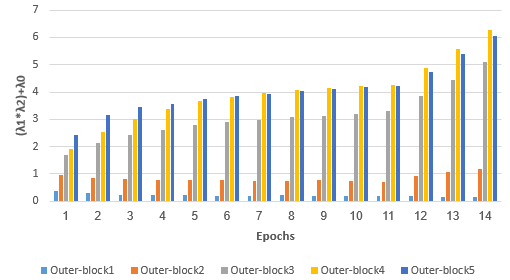}%
\caption{The values of total skip-connections for each outer clock, i.e. the amount of input that skips each outer-block through the constant highway gates, as a function of the training epochs. }
\label{fig:lambda}
\end{figure*}

Fig.~\ref{fig:lambda} depicts the progression of these values by epochs for our five outer-blocks description network. One can observe that when the network is fully trained after epoch 14, the values of the deeper outer-blocks are higher than the shallow outer-blocks. This means that the low-level blocks are hardly skipped, while the information in the fully trained network tends to skip the upper blocks more. 
This phenomenon is increasing with the training epochs: as training progresses, the skip values of the high-level blocks are increasing and the values at the shallower layers are decreasing.

\section{Reflective confidence}
\label{ap:conf}
\begin{figure*}[t]
\centering
  \includegraphics[width=0.7\linewidth]{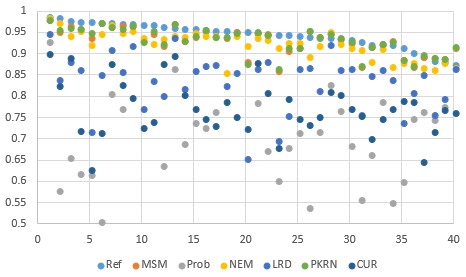}
\caption{AUC of confidence measures on 40 random validation images from the KITTI 2012 stereo data sets, ordered by the reflective confidence score. The reflective confidence (Ref) is shown to outperform the different measures, described in Sec.~\ref{ablation}, on almost every image.}
\label{fig:kitti2012_conf}
\end{figure*}

Similar to Fig~\ref{fig:kitti2015_conf}, in Fig~\ref{fig:kitti2012_conf} that depicts the comparison of the different measures described in Sec~\ref{ablation} and our reflective confidence on 40 random validation images from the KITTI 2012 dataset, one can see that our reflective confidence measure performs better on almost every image.

\section{Runtime}

We measure the runtime required for the disparity map computation of an image pair, on a computer with a single NVIDIA Titan X (Pascal) graphics processor unit. The images are taken from the KITTI 2012 data set and their sizes are $1242\times 350$, with 228 possible disparities. 

Tab.~\ref{tab:runtime} presents the measured runtime of a single computation of each step in our accurate and fast pipelines, as well as the number of iterations required for the full prediction. The three main differences between the pipelines are: (i) The fast description sub-network contains four outer blocks and the accurate five. (ii) The fast decision sub-network contains the dot product between the two image descriptors, as described in Sec. 3 of the paper. (iii) No cost aggregation is performed in the fast pipeline. 

The total runtime is 48 seconds for our accurate method and 2.84 seconds for the fast. However, this can be much reduced by parallel computation of the two bottlenecks.
The first is the computation time of the left and right image descriptors, which is 40 percent of the fast method's runtime, and can be reduced by computing the two descriptors in parallel. The second is the time of $disparity\_max$ forward passes in the decision sub-network, that are required in order to compute the matching cost for every possible disparity. Each forward pass can be computed in parallel, and thus reduce up to 90 percent of the accurate method's runtime and another 11 percent of the fast method's runtime.

\begin{table*}[t]
\centering

\begin{tabular}{lccc|ccc}
                         & \multicolumn{3}{c}{Fast}     	& \multicolumn{3}{c}{Accurate} \\ \cline{2-4} \cline{5-7} 
Component                & Runtime & Iterations & Total 	& Runtime & Iterations  & Total \\ \hline
Description sub-network  & 0.61    & 2          & 1.22  	& 0.90    & 2           & 1.8      \\
Decision sub-network     & 0.0007  & 456        & 0.31  	& 0.097  & 456         & 44.23      \\
CBCA                     &  -      & 0          & 0     	& 0.08   & 4           & 0.32       \\
SGM                      &  0.12   & 2          & 0.24      & 0.12   & 2			& 0.24      \\
Global disparity network &  1.04   & 1          & 1.04      & 1.04   & 1          & 1.04      \\
Outlier interpolation    &  0.0008 & 1          & 0.0008    & 0.0008 & 1          & 0.0008      \\
Sub-pixel enhancement    &  0.0001 & 1          & 0.0001    & 0.0001 & 1          & 0.0001      \\
Smoothing and refinement &  0.01   & 1          & 0.01      & 0.01   & 1         & 0.01      \\ \hdashline
Everything else          &  0.02   & -          & 0.02      & 0.02   & -          & 0.02      \\ \hline
\bf{Total}               &         &         & \bf{2.84s}   &         &             & \bf{47.93s}     
\end{tabular}
\caption{The runtime in seconds that is required for prediction of each component on a single NVIDIA Titan X (Pascal). In order to compute the matching cost map on the KITTI data set, the description network has to be run twice: once to create the left image descriptors and once to create the right image descriptors. The decision network has to be run $disparity\_max=228$ times, once for every possible disparity. In order to perform the left-right consistency check in the outlier interpolation step, the matching cost map is computed twice: once when using the left image as a reference image and once when using the right image as reference. The two image descriptors can be reused, but another $disparity\_max$ forward passes in the decision network are required, followed by CBCA and SGM computations.}
\label{tab:runtime}
\end{table*}

\end{document}